\documentclass{article}

\usepackage{PRIMEarxiv}
\usepackage{microtype}
\usepackage{graphicx}
\usepackage{float}
\usepackage{subfigure}
\usepackage{stfloats}
\usepackage{booktabs} 
\usepackage{algorithm}
\usepackage{algorithmic}

\usepackage{hyperref}

\usepackage{amsmath}
\usepackage{amssymb}
\usepackage{mathtools}
\usepackage{amsthm}
\usepackage{array}

\usepackage{caption}

\usepackage{hyperref}
\usepackage[capitalize,noabbrev]{cleveref}

\theoremstyle{plain}
\newtheorem{theorem}{Theorem}[section]

\newtheorem{lemma}[theorem]{Lemma}

\theoremstyle{definition}

\theoremstyle{remark}

\usepackage[utf8]{inputenc} 
\usepackage[T1]{fontenc}    
\usepackage{hyperref}       
\usepackage{url}            
\usepackage{booktabs}       
\usepackage{amsfonts}       
\usepackage{nicefrac}       
\usepackage{microtype}      
\usepackage{lipsum}
\usepackage{fancyhdr}       
\usepackage{graphicx}       
\graphicspath{{media/}}     

\title{Function Fitting Based on Kolmogorov-Arnold Theorem and Kernel Functions
}

\author{
  Jianpeng Liu \\
  XJTLU \\
  \texttt{Jianpeng.Liu23@student.xjtlu.edu.cn} \\
   \And
  Qizhi Pan \\
  XJTLU \\
  \texttt{Qizhi.Pan23@student.xjtlu.edu.cn} \\
}

\begin{document}
\maketitle

\begin{abstract}
This paper proposes a unified theoretical framework based on the Kolmogorov-Arnold representation theorem and kernel methods. By analyzing the mathematical relationship among kernels, B-spline basis functions in Kolmogorov-Arnold Networks (KANs) and the inner product operation in self-attention mechanisms, we establish a kernel-based feature fitting framework that unifies the two models as linear combinations of kernel functions. Under this framework, we propose a low-rank Pseudo-Multi-Head Self-Attention module (Pseudo-MHSA), which reduces the parameter count of traditional MHSA by nearly 50\%. Furthermore, we design a Gaussian kernel multi-head self-attention variant (Gaussian-MHSA) to validate the effectiveness of nonlinear kernel functions in feature extraction. Experiments on the CIFAR-10 dataset demonstrate that Pseudo-MHSA model achieves performance comparable to the ViT model of the same dimensionality under the MAE framework and visualization analysis reveals their similarity of multi-head distribution patterns. Our code is publicly available~\footnote{Code: \url{https://github.com/cfrslyr/Experiments}}.
\end{abstract}


\section{Introduction}
\label{sec:introduction}
Deep neural networks \cite{LeCun2015} have become the cornerstone of modern machine learning due to their ability to model complex data distributions and learn hierarchical features. In recent years, Kolmogorov-Arnold Networks (KANs) \cite{liu2025kankolmogorovarnoldnetworks}, based on the Kolmogorov-Arnold representation theorem\cite{kolmogorov1957}, have introduced a novel perspective for the development of neural network architectures. Unlike traditional networks that rely on static activation functions (e.g., ReLU\cite{Nair2010} or Sigmoid\cite{Britannica2023}), KANs dynamically learn activation functions by incorporating B-spline bases function, offering a flexible mechanism to capture complex and nonlinear patterns in data.\\

Meanwhile, the self-attention mechanism \cite{Vaswani2017}, as the core component of the Transformer architecture, has demonstrated exceptional performance in natural language processing and computer vision. Self-attention enables each element in a sequence to attend to all other elements, effectively capturing long-range dependencies. Its essence lies in computing attention scores via dot products, followed by scaling, Softmax normalization, and weighted summation to generate context-aware representations. The multi-head self-attention mechanism further extends this process by executing it in parallel across multiple subspaces, allowing the model to capture relationships from diverse perspectives. Notably, from a theoretical standpoint, the inner product operation in self-attention aligns with kernel methods for computing similarity in high-dimensional spaces, providing a theoretical foundation for reinterpreting attention mechanisms.\\

A deeper analysis of the implementation details of KANs and self-attention reveals intriguing mathematical connections. The output of KANs is a linear combination of B-spline basis functions\cite{deBoor1972}, where these basis functions can be viewed as kernel functions between inputs and grid points. Similarly, the inner product operation in self-attention can be interpreted as a kernel function computation, implying that attention maps are effectively linear combinations of multiple kernel functions. Inspired by this, we propose a latent feature fitting framework that replaces the traditional function construction in the Kolmogorov-Arnold representation theorem (also known as the superposition theorem) with linear combinations of kernel functions. This leads to the introduction of our superposition kernel method, which organically integrate the strengths of KANs and Transformers, offering a flexible and theoretically grounded model for feature extraction and representation learning.\\

In exploring the multi-head adaptation of the superposition kernel model for self-attention, we hypothesize that low-rank approximation methods could reduce the parameter count of multi-head self-attention while improving computational efficiency. Guided by this idea, we design Pseudo-MHSA, a variant that leverages low-rank approximation to significantly reduce parameters and computational costs while maintaining---or even enhancing---the model's expressive power. Experimental results demonstrate that the low-rank-optimized model not only approximates self-attention behavior on the CIFAR-10 \cite{Krizhevsky2009} dataset but also exhibits superior parameter efficiency and comparable performance to the standard Vision transformer (ViT) \cite{Dosovitskiy2020} baseline.\\

\subsection{Contributions}
The main contributions of this work are summarized as follows:  
\begin{itemize}  
\item Based on the Kolmogorov-Arnold representation theorem, we establish a unified kernel method framework that provides a novel theoretical interpretation of self-attention mechanisms and elucidates their convolutional properties under this framework.  
\item Guided by this method, we propose several variants of multi-head self-attention modules using different kernel functions and low-rank approximations, validating the theoretical insights through empirical evaluations.
\item Through experiments on CIFAR-10 under the Masked autoencoder (MAE) \cite{He2021} framework, we verify the effectiveness of our proposed model in feature learning and approximating self-attention behavior. The results demonstrate that the optimized model achieves advantages in both parameter efficiency and performance.
\end{itemize}

The organizational structure of the remainder of this paper is as follows: chapter \ref{sec:background} establishes the theoretical foundation, discussing the Kolmogorov-Arnold representation theorem and its integration with kernel methods. chapter \ref{sec:architecture} introduces the proposed model architecture. chapter \ref{experiment} presents experimental results. Finally, the conclusion outlines future research directions and potential extensions.\\

\section{Theoretical Background}
\label{sec:background}
This section establishes the theoretical foundation for our approach. We first present the Kolmogorov-Arnold representation theorem in Section~\ref{sec:kolmogorov-arnold}, which forms the mathematical basis for our model formulation. Subsequently, Section~\ref{sec:self-attention} demonstrates how the self-attention mechanism can be unified within this theoretical framework.\\

\subsection{Kolmogorov-Arnold Representation Theorem}
\label{sec:kolmogorov-arnold}

The Kolmogorov-Arnold representation theorem provides fundamental insights into the expressive power of neural networks through its decompositional perspective of multivariate functions.
\begin{theorem}[Kolmogorov-Arnold]
For any continuous multivariate function \( f(\mathbf{x}): \mathbb{R}^D \to \mathbb{R} \) defined on a bounded domain, there exist:
\begin{itemize}
    \item Outer functions: \( \{\phi_h\}_{h=1}^H \) where \( H = 2D+1 \)
    \item Inner functions: \( \{\psi_{h,d}\}_{h=1,d=1}^{H,D} \)
\end{itemize}
such that:
\begin{equation}
    f(\mathbf{x}) = \sum_{h=1}^{H} \phi_{h}\left( \sum_{d=1}^D \psi_{h,d}(x_d) \right)
\end{equation}
where all \(\phi_h\) and \(\psi_{h,d}\) are continuous univariate functions.
\end{theorem}
The construction of these functions finds practical realization through Kolmogorov-Arnold Networks (KANs), where both inner and outer functions are implemented as linear combinations of B-spline basis functions:

\begin{align}
    \psi_{h,d}(x_d) &= \sum_{i=1}^N w_i^{(h,d)} B_i(x_d) \label{eq:inner-spline} \\
    \phi_h(\psi) &= \sum_{j=1}^M w_j^{(h)} B'_j(\psi) \label{eq:outer-spline}
\end{align}

where \( B_i(\cdot) \) denotes B-spline basis functions of degree \( k \). These bases possess deep connections to fundamental solutions of linear differential operators. Specifically, degree-\(k \) B-splines can be expressed as the differences of Green's function for the differential operator \( L = D^{k+1} \), while Green's functions can directly induce the kernel function of the differential operator $L$ \cite{zhang2008}. By exchanging the calculation order we can get:

\begin{equation}
    B_i(x) = \sum_{j=1}^R b(x, s_j)
\end{equation}

where \( \{s_j\}_{j=1}^R \) represents the spline knot positions. This Green's function perspective naturally leads to kernel-based interpretations. For a multivariate function \( g: \mathbb{R}^D \to \mathbb{R} \), we consider approximations using kernel expansions:

\begin{equation}
    g(\mathbf{x}) = \sum_{r=1}^R \sum_{d,=1}^D \sum_{d'=1}^D w_{r,d,d'} k(x_d, s_{r,d'})
\end{equation}

where \( \mathbf{S} = (s_{r,d'}) \in \mathbb{R}^{R \times D} \) is a reference matrix. Defining the kernel matrix between input \( \mathbf{x} \) and reference vector \( \mathbf{s}_r \) as:

\begin{equation}
    \mathbf{K}(\mathbf{x}, \mathbf{s}_r) = 
    \begin{bmatrix}
        k(x_1, s_{r,1}) & \cdots & k(x_1, s_{r,D}) \\
        \vdots & \ddots & \vdots \\
        k(x_D, s_{r,1}) & \cdots & k(x_D, s_{r,D})
    \end{bmatrix}
\end{equation}

we can express the function approximation using Frobenius inner products:

\begin{equation}
    g(\mathbf{x}) = \sum_{r=1}^R \langle \mathbf{W}_r, \mathbf{K}(\mathbf{x}, \mathbf{s}_r) \rangle_F
\end{equation}

This formulation aligns with the Kolmogorov-Arnold decomposition when we consider:

\begin{equation}
    \psi_{h,d}(\cdot) := \sum_{r=1}^R \sum_{d=1}^D \sum_{d'=1}^D w_{h,r,d,d'} k(\cdot, s_{r,d'})
\end{equation}

Letting \( \Psi(\mathbf{x}) = (\psi_1(\mathbf{x}), ..., \psi_H(\mathbf{x})) \) where \( \psi_h = \sum_{d=1}^D \psi_{h,d}(x_d) \), and \( \Phi(\Psi) = \sum_{h=1}^H \phi_h(\psi_h) \),

\begin{align}
    f(\mathbf{x}) &= \Phi \circ \Psi(\mathbf{x}) \\
    \Psi(\mathbf{x}) &= \left( \sum_{r=1}^R \langle \mathbf{W}_{h,r}, \mathbf{K}(\mathbf{x}, \mathbf{s}_r) \rangle_F \right)_{h=1}^H \\
    \Phi(\mathbf{z}) &= \left( \sum_{r'=1}^{R'} \langle \mathbf{W}'_{e,r'}, \mathbf{K}'(\mathbf{z}, \mathbf{s}'_{r'}) \rangle_F \right)_{e=1}^E
\end{align}

where:
\begin{itemize}
    \item \( \mathbf{W}_{h,r} \in \mathbb{R}^{D \times D} \): Inner function parameters
    \item \( \mathbf{W}'_{e,r'} \in \mathbb{R}^{H \times H} \): Outer function parameters
    \item \( \mathbf{S} \in \mathbb{R}^{R \times D}, \mathbf{S}' \in \mathbb{R}^{R' \times H} \): Reference matrices
\end{itemize}

For batched sequential inputs $\mathbf{X} \in \mathbb{R}^{B \times S \times D}$ and batched reference $\mathbf{Ref} \in \mathbb{R}^{B \times R \times D}$, we define the kernel tensor $\mathbf{K}(\mathbf{X}, \mathbf{Ref}) \in \mathbb{R}^{B \times S \times R \times D \times D}$:
\begin{equation}
    \mathbf{K}(\mathbf{X}, \mathbf{Ref})_{bsrd_1d_2} = k(\mathbf{X}_{bsd_1}, \mathbf{Ref}_{brd_2}).
\end{equation}
Using the kernel tensor and Einstein summation to simplify the calculation, we obtain our complete model formulation: 

\begin{lemma}[Superposition-Kernel Formulation]
\label{lemma:KA-Kernel}
For batched inputs $\mathbf{X} \in \mathbb{R}^{B \times S \times D}$ and mapping function $\mathbf{f}(\mathbf{X}): \mathbb{R}^{B \times S \times D} \to \mathbb{R}^{B \times S \times E}$, the function approximation admits $\mathbf{f} = \Phi \circ \Psi$, and:
\begin{align}
    \Psi(\mathbf{X})_{bsh} &= \mathbf{K}^{inner}(\mathbf{X}, \mathbf{Ref}^{inner})_{bsrd_1d_2} \mathbf{W}_{hrd_1d_2}^{inner} \\
    \Phi(\Psi)_{bse} &= \mathbf{K}^{outer}(\Psi, \mathbf{Ref}^{outer})_{bsrh_1h_2} \mathbf{W}_{erh_1h_2}^{outer}
\end{align}
where,
\begin{itemize}
    \item \( \mathbf{W}^{inner} \in \mathbb{R}^{H \times R \times D \times D} \), \( \mathbf{W}^{outer} \in \mathbb{R}^{E \times R' \times H \times H} \): Parameters of the inner function $\Psi$ and outer function $\Phi$.
    \item \( \mathbf{Ref}^{inner} \in \mathbb{R}^{B \times R \times D}, \mathbf{Ref}^{outer} \in \mathbb{R}^{B \times R' \times H} \): Reference matrices of the inner kernel tensor $\mathbf{K}^{(inner)}$ and outer kernel tensor $\mathbf{K}^{(outer)}$
\end{itemize}
\end{lemma}

\begin{figure}[htpb]
    \centering
    \includegraphics[width=\linewidth]{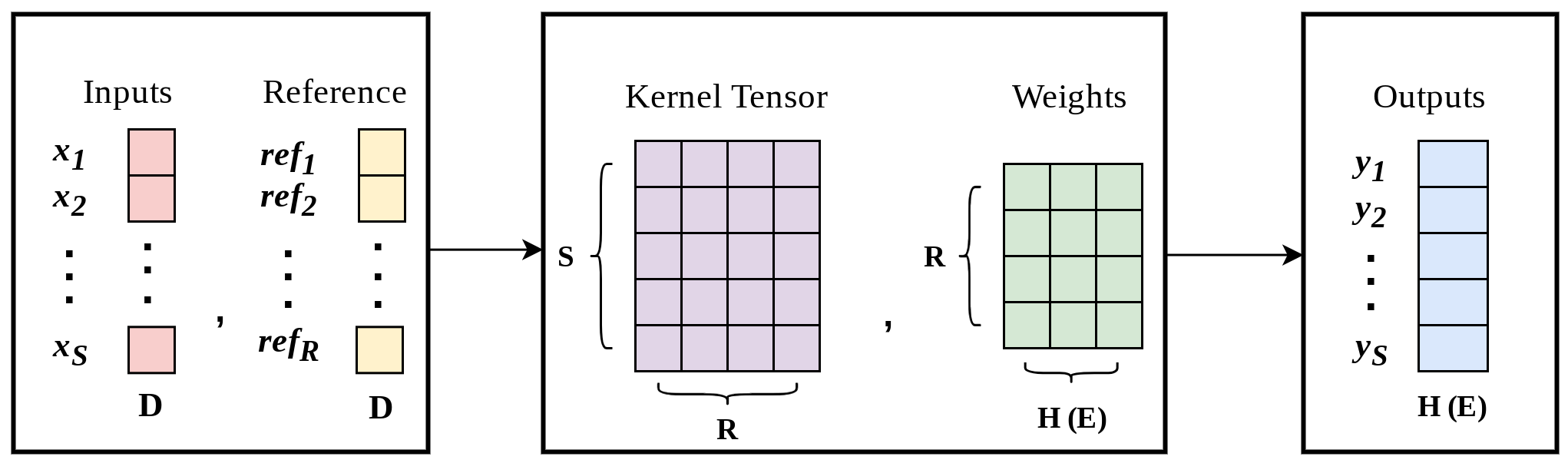}
    \caption{inner/outer function progress. In the left block, each cell is a vector of dimension D, and ref matrix can be trainable parameters or fixed matrix like inputs itself. In the middle block, each cell is a D$\times$D matrix. For kernel tensor, the cell at s-th row and h-th column, noted as $\textbf{K}_{sr}$, is computed by $x_{s}$ and $ref_{r}$. Using $\textbf{W}_{hr}$ to indicate the cell at h-th row and r-th column, the output $y_{sh}$ is calculated as lemma (\ref{lemma:KA-Kernel})
    }
    \label{fig:kernelmatrix}
\end{figure}

The mapping function $\mathbf{f}$ thus becomes a sequence of tensor contractions that preserve the Kolmogorov-Arnold structure as shown in Fig~\ref{fig:kernelmatrix}.

\subsection{Self-Attention Mechanism} 
\label{sec:self-attention}

The standard self-attention mechanism emerges as a special case of our kernel-based framework. Consider an input sequence \( \mathbf{X} \in \mathbb{R}^{S \times D} \) with \( S \) tokens of dimension \( D \). Let \( \mathbf{W}^q, \mathbf{W}^k, \mathbf{W}^v \in \mathbb{R}^{D \times E} \) denote the query, key, and value projection matrices respectively. The attention computation (excluding softmax normalization) can be reformulated as:

\begin{equation}
    \mathbf{Y} = \mathbf{X}\mathbf{W}^q (\mathbf{W}^k)^\top \mathbf{X}^\top \mathbf{X}\mathbf{W}^v
\end{equation}

For individual token representations \( \mathbf{x}_i \in \mathbb{R}^D \), this operation decomposes into:

\begin{equation}
\label{attn calcu}
    f(\mathbf{x}_i) = \mathbf{x}_i \mathbf{W}_{attn} \mathbf{X}^\top \mathbf{X} \mathbf{W}^v, \quad \text{where} \quad \mathbf{W}_{attn} = \mathbf{W}^q (\mathbf{W}^k)^\top
\end{equation}

This formulation reveals an elegant correspondence with our kernel framework through the following identifications:

\begin{enumerate}
    \item \textbf{Inner Function}: The feature interaction layer implements:
    \begin{equation}
    \label{eq:sa-inner}
        a_h = \sum_{r=1}^R \langle \mathbf{W}_{h,r}, \mathbf{K}(\mathbf{x}_i, \mathbf{x}_h) \rangle_F
    \end{equation}
    with linear kernel \( \mathbf{K}(\mathbf{x}, \mathbf{y}) = \mathbf{x}\mathbf{y}^\top \), reference matrix \( \mathbf{Ref}^{(inner)} = \mathbf{X} \), and parameter tensor \( \mathbf{W}_{h,r} = \delta_{h,r}\mathbf{W}_{attn} \) where \( H = R \).

    \item \textbf{Outer Function}: The context aggregation layer implements: 
    \begin{equation}
    \label{eq:sa-outer}
        y_e = \sum_{r'=1}^D \langle \mathbf{W}'_{e,r'}, \mathbf{K}(\mathbf{a}, \mathbf{x}'_{r'}) \rangle_F
    \end{equation}
    with \( \mathbf{K}(\mathbf{x}, \mathbf{y}) = \mathbf{x}\mathbf{y}^\top \), \( \mathbf{Ref}^{(outer)} = \mathbf{X}^\top \), and \( \mathbf{W}'_{e,r'} = w_{e,r'}^{v}\mathbf{I}_S \) where \( R'=D \).
\end{enumerate}

This analysis establishes that standard self-attention mechanisms can be viewed as linear combinations of kernel operators.

\subsubsection{Convolutional Interpretation}
\label{ssec:conv-attention}

\begin{figure}[H]
    \centering
    \includegraphics[width=0.8\linewidth]{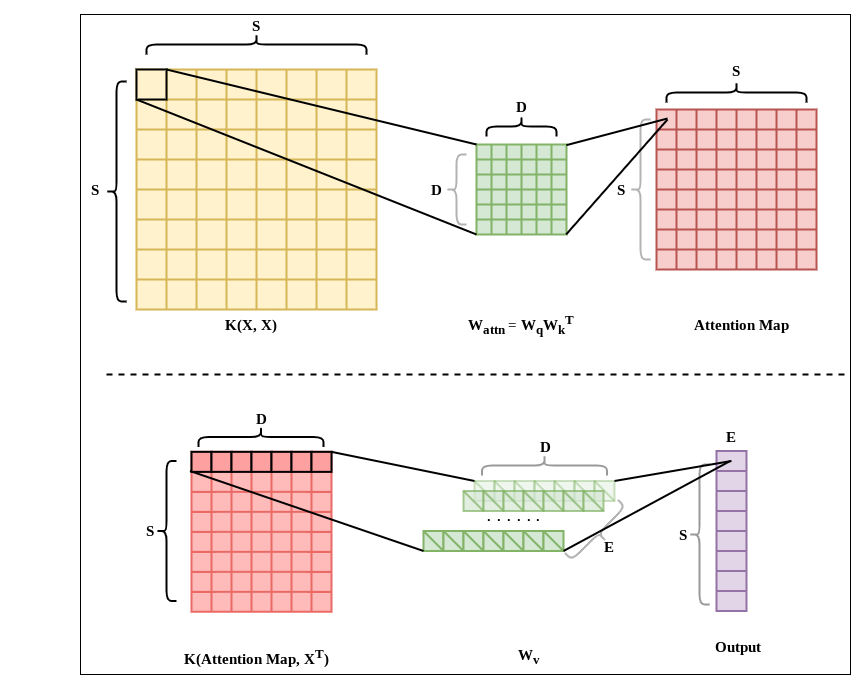}
    \caption{%
    Architectural equivalence between self-attention and convolutional operations.
    (a) \textbf{Inner Convolution}: Each $D \times D$ block in $\mathbf{K}(\mathbf{X}, \mathbf{X})$ (reshaped accordingly) undergoes depthwise convolution with the kernel $\mathbf{W}_{\mathrm{attn}}$.
    (b) \textbf{Outer Convolution}: For output channel $e$, the outer kernel tensor $\mathbf{K}(\textbf{Attention Map}, \mathbf{X}^\top)$ (reshaped) is processed through strided convolutions with a set of concatenated identity-mapped kernels 
    $\bigl[w_{e,1} I_S, \dots, w_{e,d} I_S\bigr]$ 
    to generate the final output.}
    \label{fig:attn-conv-equivalence}
\end{figure}

The convolutional property of self-attention was established by Cordonnier, J. B., Loukas, A. and Jaggi, M.\cite{cordonnier2020relationshipselfattentionconvolutionallayers}. Building on this, we provide a kernel perspective that further enables the convolutional interpretation of the Transformer, as shown in Fig.\ref{fig:attn-conv-equivalence}. For batched input \( \mathbf{X} \in \mathbb{R}^{B \times S \times D} \), consider the inner kernel tensor \( \mathbf{K}^{inner}(\mathbf{X}, \mathbf{X}) \) constructed with linear kernels \( k(x, y) = xy \). When reshaped to \( \mathbf{K}^* \in \mathbb{R}^{B \times (SD) \times (RD)} \), the inner function in equation~\eqref{eq:sa-inner} becomes equivalent to a 2-dimensional convolution:

\begin{equation}
    \Psi(\mathbf{X}) = \text{Conv}_{D \times D}(\mathbf{K}^*, \mathbf{W}^{inner})
\end{equation}

where: 
\begin{itemize}
    \item \( \mathbf{W}^{inner} = \mathbf{W}_{attn} \) functions as a convolutional filter
    \item The subscript \( D \times D \) denotes convolution strides
    \item Input/output channels of the filter are both 1.
\end{itemize}

The outer function in Equation~\eqref{eq:sa-outer} admits a similar interpretation through strided convolutions over the outer kernel tensor.

This unified perspective grounded in the Kolmogorov-Arnold theorem provides new insights into the representational capabilities of attention mechanism, while maintaining full compatibility with standard deep learning primitives. The mathematical synthesis presented here opens avenues for developing hybrid architectures that combine the strengths of both paradigms.

\section{Model Architecture}
\label{sec:architecture}

Building upon the theoretical framework established in Section~\ref{sec:background}, we now present our model architecture. While Lemma~\ref{lemma:KA-Kernel} provides a general functional approximation scheme, direct implementation faces critical computational bottlenecks. Specifically, for a batch input $\mathbf{X} \in \mathbb{R}^{B \times S \times D}$ with self-referential computation, the corresponding kernel tensor $\mathbf{K} \in \mathbb{R}^{B \times S \times S \times D \times D}$ becomes computationally intractable. As a concrete example, with standard float32 precision and typical dimensions ($B=100$, $S=64$, $D=256$), directly storing the kernel tensor would require over 100GB of GPU memory. While alternative computation strategies without full tensor storage exist, naive implementations using loops would lead to impractical training durations. This fundamental challenge motivates our adaptation of the multi-head attention mechanism (MHSA). During our exploration, we developed a low-rank approximation approach from a kernel perspective, which reduces the parameters of the self-attention layer by nearly half. This method can be regarded as an optimization strategy for the self-attention mechanism, offering a more efficient implementation while preserving its expressive power.

\subsection{Multi-Head Self-Attention}
\label{ssec:mhsa}

We first revisit the mathematical formulation of multi-head attention. Given an input sequence $\mathbf{X} \in \mathbb{R}^{S \times D}$ with $n$ attention heads and $D_{\text{head}} = D/n$ (assuming divisibility), let $\mathbf{W}^q$, $\mathbf{W}^k$, $\mathbf{W}^v$, and $\mathbf{W}^O$ denote the query, key, value, and output projection matrices, respectively. Ignoring softmax normalization and bias terms for simplicity, we have:
\begin{gather}
    \mathbf{Q} = \mathbf{X} \mathbf{W}^q = [\mathbf{Q}_1, \dots, \mathbf{Q}_n], \\
    \mathbf{K} = \mathbf{X} \mathbf{W}^k = [\mathbf{K}_1, \dots, \mathbf{K}_n], \\
    \mathbf{V} = \mathbf{X} \mathbf{W}^v = [\mathbf{V}_1, \dots, \mathbf{V}_n].
\end{gather}
For the $i$-th head:
\begin{equation}
    \text{Head}_i = \mathbf{Q}_i \mathbf{K}_i^\top \mathbf{V}_i.
\end{equation}
The final output is computed as:
\begin{align}
\label{eq:mhsa-output}
    \text{Output} &= \big[ \text{Head}_1, \dots, \text{Head}_n \big] \mathbf{W}^O \\
           &= \sum_{i=1}^n \text{Head}_i \mathbf{W}_i^O \\
           &= \sum_{i=1}^n \mathbf{X} \mathbf{W}_i^q (\mathbf{W}_i^k)^\top \mathbf{X}^\top \mathbf{X} \mathbf{W}_i^v \mathbf{W}_i^O,
\end{align}
where $\mathbf{W}_i^O \in \mathbb{R}^{D_{\text{head}} \times D}$ is the $i$-th split of $\mathbf{W}^O$:
\begin{equation}
    \mathbf{W}^O = \begin{bmatrix}
                \mathbf{W}_1^O \\ \vdots \\ \mathbf{W}_n^O
          \end{bmatrix}.
\end{equation}

\subsubsection{Low-Rank Approximation}
\label{sssec:low-rank}

Following the representation in Equation~\eqref{attn calcu}, $\text{Head}_i$ can be expressed as:
\begin{equation}
    \text{Head}_i = \mathbf{X} \mathbf{W}_i^{(\text{attn})} \mathbf{X}^\top, \quad \text{where} \quad \mathbf{W}_i^{(\text{attn})} = \mathbf{W}_i^q (\mathbf{W}_i^k)^\top.
\end{equation}
Noting that $\mathbf{W}_i^{(\text{attn})}$ has a rank bounded by $D_{\text{head}}$, this low-rank structure motivates our approximation scheme (which, to some extent,  approximates the SVD decomposition \cite{Golub1996}):
\begin{gather}
    \mathbf{W}_i^{(\text{attn})} \approx \mathbf{U}_i \mathbf{A}_i \mathbf{U}_i^\top, \\
    \mathbf{I}_E \approx \mathbf{U}_i \mathbf{U}_i^\top,
\end{gather}
where $\mathbf{U}_i \in \mathbb{R}^{D \times D_{\text{head}}}$ are column-wise orthonormal, and $\mathbf{A}_i \in \mathbb{R}^{D_{\text{head}} \times D_{\text{head}}}$. The $i$-th head output is then approximated as:
\begin{align}
    \text{Head}_i &\approx \mathbf{X} \mathbf{U}_i \mathbf{A}_i \mathbf{U}_i^\top \mathbf{X}^\top \mathbf{X} \mathbf{U}_i \mathbf{U}_i^\top \mathbf{W}_i^v \\
            &= \widetilde{\mathbf{X}}_i \mathbf{A}_i \widetilde{\mathbf{X}}_i^\top \widetilde{\mathbf{X}}_i \widetilde{\mathbf{W}}_i^v,
\end{align}
where $\widetilde{\mathbf{X}}_i = \mathbf{X} \mathbf{U}_i$ is the $i$-th split of $\mathbf{X} \mathbf{U} = [\mathbf{X} \mathbf{U}_1, \dots, \mathbf{X} \mathbf{U}_n]$ with $\mathbf{U}$ being a $D \times D$ projection matrix, and $\widetilde{\mathbf{W}}_i^v$ acting as a $D_{\text{head}} \times D_{\text{head}}$ transformation. 

Considering the output shown in Equation~\eqref{eq:mhsa-output}, we can derive:
\begin{align}
\label{eq: attn output}
    \text{Output} &= \sum_{i=1}^{n} \text{Head}_i \mathbf{W}_i^O \\
        &\approx \sum_{i=1}^{n} \widetilde{\mathbf{X}}_i \mathbf{A}_i \widetilde{\mathbf{X}}_i^\top \widetilde{\mathbf{X}}_i \widetilde{\mathbf{W}}_i^v \mathbf{W}_i^O \\
        &= \sum_{i=1}^{n} \widetilde{\mathbf{X}}_i \mathbf{A}_i \widetilde{\mathbf{X}}_i^\top \widetilde{\mathbf{X}}_i \mathbf{P}_i,
\end{align}
where $\mathbf{P}_i = \widetilde{\mathbf{W}}_i^v \mathbf{W}_i^O \in \mathbb{R}^{D_{\text{head}} \times D}$. This analysis reveals that an efficient implementation of multi-head self-attention can be achieved using just three core tensors:
\begin{enumerate}
    \item $\mathbf{U} \in \mathbb{R}^{D \times D}$: Input projection matrix
    \item $\mathbf{A} \in \mathbb{R}^{n \times D_{\text{head}} \times D_{\text{head}}}$: Multi-head attention tensor
    \item $\mathbf{P} \in \mathbb{R}^{D \times D}$: Output projection matrix
\end{enumerate}

We present the following pseudo-implementation for multi-head self-attention:

\begin{algorithm}[H]
\caption{Pseudo-MHSA Block}
\label{alg:pseudo-mhsa}
\begin{algorithmic}[1]
\STATE \textbf{Input:}
\STATE \quad Data: $\mathbf{X}$: ($B$, $S$, $D$)
\STATE \quad Number of heads: $n$, $D_{\text{head}} = D // n$
\STATE \textbf{Parameters:}
\STATE \quad In-projection parameters: $\mathbf{U}$: ($D$, $D$), $\mathbf{b}_1$: ($D$)
\STATE \quad Attention weights: $\mathbf{A}$: ($n$, $D_{\text{head}}$, $D_{\text{head}}$)
\STATE \quad Out-projection parameter: $\mathbf{P}$: ($D$, $D$), $\mathbf{b}_2$: ($D$)

\STATE \textbf{Compute the In-projection:} 
\STATE \quad $\widetilde{\mathbf{X}} = \mathbf{X} \mathbf{U} + \mathbf{b}_1$
\STATE \quad Split $\widetilde{\mathbf{X}}$ into $[\widetilde{\mathbf{X}}_0, \dots, \widetilde{\mathbf{X}}_{n-1}]$

\STATE \textbf{Compute the self-attention map:} 
\FOR{$i=0$ \TO $n-1$}
    \STATE $\mathbf{AttnMap}_i = \widetilde{\mathbf{X}}_i \mathbf{A}_i \widetilde{\mathbf{X}}_i^\top$
    \STATE $\mathbf{AttnMap}_i = \text{Softmax}\!\left(\dfrac{\mathbf{AttnMap}_i}{D_{\text{head}}}\right)$
\ENDFOR

\STATE \textbf{Compute the self-attention output:} 
\FOR{$i=0$ \TO $n-1$}
    \STATE $\mathbf{Head}_i = \mathbf{AttnMap}_i \widetilde{\mathbf{X}}_i$
\ENDFOR
\STATE $\mathbf{Head} = \textbf{Concat}[\mathbf{Head}_0, \dots, \mathbf{Head}_{n-1}]$

\STATE \textbf{Compute the Out-projection:} 
\STATE \quad $\mathbf{Output} = \mathbf{Head} \mathbf{P} + \mathbf{b}_2$
\STATE \textbf{return} $\mathbf{Output}$
\end{algorithmic}
\end{algorithm}

\subsubsection{Gaussian-MHSA}
\label{sssec:gaussian-mhsa}

Our Gaussian-MHSA Block variant integrates the function fitting framework from Lemma~\ref{lemma:KA-Kernel} through three key components as illustrated in Algorithm~\ref{alg:pseudo-mhsa}: multi-head In-projection, attention mapping, and Out-projection. 

For the $i$-th head of the In-projection output $\widetilde{\mathbf{X}}^i$, we define:
\begin{align}
    \Psi(\widetilde{\mathbf{X}}^i)_{bsr} &= \mathbf{AttnMap}(\widetilde{\mathbf{X}}^i) \\
    &\triangleq \mathbf{K}(\widetilde{\mathbf{X}}^i, \widetilde{\mathbf{X}}^i)_{bsrd_1d_2} \mathbf{A}^i_{d_1d_2}, \\
    \Phi(\Psi_i) &= \mathbf{K}^{(\text{outer})}(\mathbf{AttnMap}_i, \widetilde{\mathbf{X}}^i) \mathbf{W}^{i},
\end{align}
where the inner kernel function is defined as $k(x, y) = e^{-\frac{\|x-y\|^2}{2\sigma^2}}$ \cite{Scholkopf2002} with hyperparameter $\sigma$, and the outer kernel function simplifies to:
\begin{equation}
    \mathbf{K}^{(\text{outer})}(\mathbf{AttnMap}_i, \widetilde{\mathbf{X}}^i) = \mathbf{AttnMap}_i \widetilde{\mathbf{X}}^i.
\end{equation}

Through the above definitions, our Gaussian-MHSA Block implementation remains fully aligned with the functional decomposition prescribed by Lemma~\ref{lemma:KA-Kernel}. In the current implementation, we adopt a non-parametric inner function by fixing $\mathbf{A}^i = \mathbf{1}_{D}$ for the Gaussian-MHSA Block, as detailed in Algorithm~\ref{alg:gaussian-mhsa}.

\begin{algorithm}[H]
\caption{Gaussian-MHSA Block}
\label{alg:gaussian-mhsa}
\begin{algorithmic}[1]
\STATE \textbf{Input:}
\STATE \quad Data: $\mathbf{X}$: ($B$, $S$, $D$)
\STATE \quad Number of heads: $n$, $D_{\text{head}} = D // n$
\STATE \textbf{Parameters:}
\STATE \quad In-projection parameters: $\mathbf{U}$: ($D$, $D$), $\mathbf{b}_1$: ($D$)
\STATE \quad Out-projection parameter: $\mathbf{P}$: ($D$, $D$), $\mathbf{b}_2$: ($D$)

\STATE \textbf{Compute the In-projection:} 
\STATE \quad $\widetilde{\mathbf{X}} = \mathbf{X} \mathbf{U} + \mathbf{b}_1$
\STATE \quad Split $\widetilde{\mathbf{X}}$ into $[\widetilde{\mathbf{X}}_0, \dots, \widetilde{\mathbf{X}}_{n-1}]$

\STATE \textbf{Compute the self-attention map:} 
\FOR{$i=0$ \TO $n-1$}
    \STATE $\mathbf{AttnMap}_i = \sum_{d_1, d_2} \mathbf{K}(\widetilde{\mathbf{X}}_i, \widetilde{\mathbf{X}}_i)$
    \STATE $\mathbf{AttnMap}_i = \text{Softmax}\!\left(\dfrac{\mathbf{AttnMap}_i}{D_{\text{head}}}\right)$
\ENDFOR

\STATE \textbf{Compute the self-attention output:} 
\FOR{$i=0$ \TO $n-1$}
    \STATE $\mathbf{Head}_i = \mathbf{AttnMap}_i \widetilde{\mathbf{X}}_i$
\ENDFOR
\STATE $\mathbf{Head} = \textbf{Concat}[\mathbf{Head}_0, \dots, \mathbf{Head}_{n-1}]$

\STATE \textbf{Compute the Out-projection:} 
\STATE \quad $\mathbf{Output} = \mathbf{Head} \mathbf{P} + \mathbf{b}_2$
\STATE \textbf{return} $\mathbf{Output}$
\end{algorithmic}
\end{algorithm}

\textbf{Scaling}: The scaling strategy adopted is to normalize by $D_{\text{head}}$ instead of the conventional $\sqrt{D_{\text{head}}}$ to better control output variance. Let $a_{sr} = (\mathbf{AttnMap})_{sr}$, then:
\begin{align}
    a_{sr} &= \sum_{d_1,d_2} k(\tilde{\mathbf{x}}_{sd_1}, \tilde{\mathbf{x}}_{rd_2}) w_{d_1d_2}.
\end{align}
Assuming layer-normalized \cite{Ba2016} projections for $\tilde{\mathbf{x}}_{s}$ and $\tilde{\mathbf{x}}_{r}$ and zero-mean initialization with variance $\sigma^2_w$ for attention parameters $w_{d_1d_2}$, with independence between kernel and attention weights and independence among $w_{d_1d_2}$, the variance develops as:
\begin{equation}
    \text{Var}(a_{sr}) = D_{\text{head}}^2 (\sigma^2_k + \mu^2_k) \sigma^2_w,
\end{equation}
where $\mu_k$ and $\sigma^2_k$ are the mean and variance of the kernel function $k(x, y)$ for normalized input $x$ and $y$.

\subsection{Model Framework}
\label{ssec:model-framework}

\begin{figure}[H]
   \centering
        \includegraphics[width=0.9\linewidth]{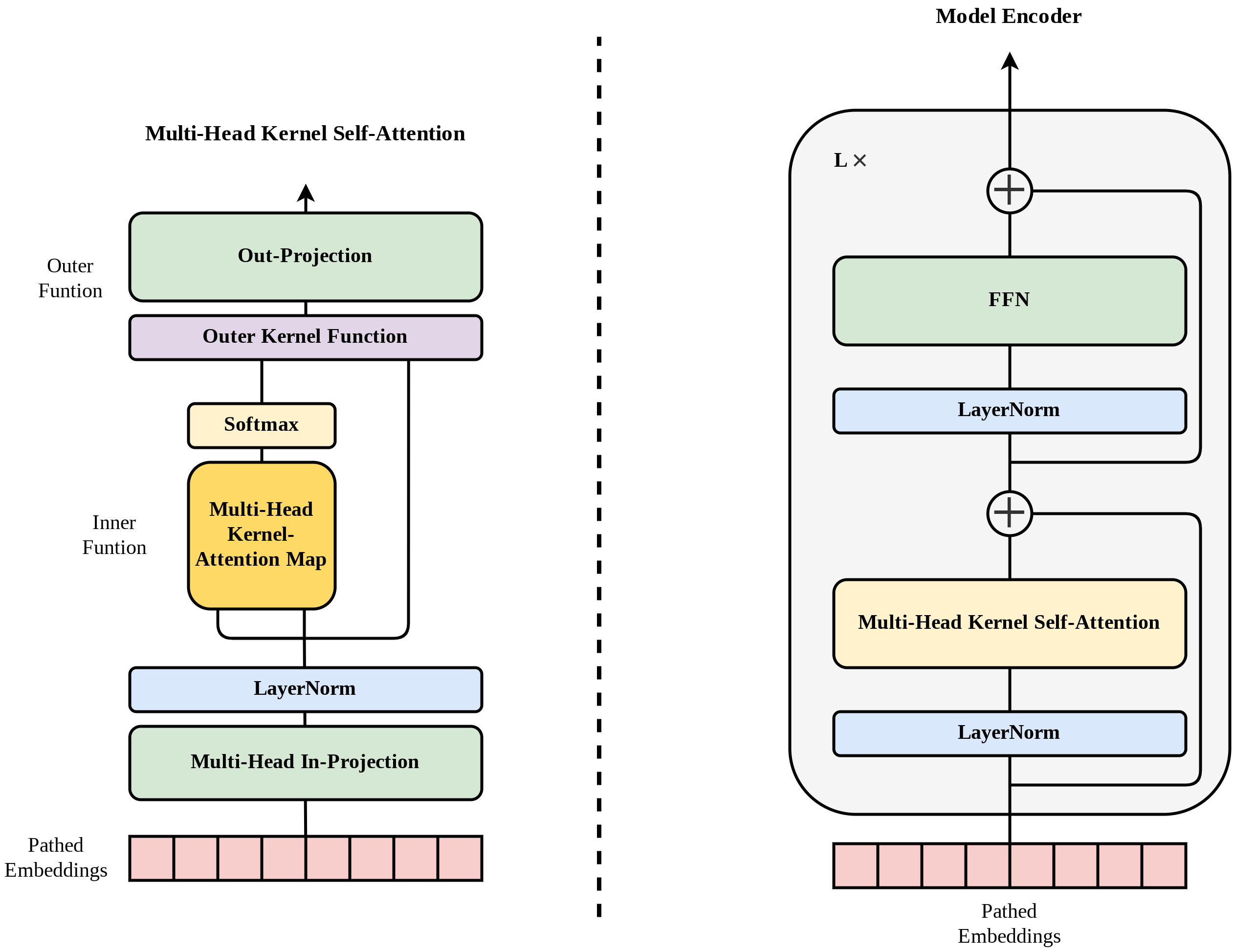}
        \caption{Attention mechanism and model encoder. The multi-head kernel self-attention mechanism comprises the In-projection step, the multi-head kernel self-attention as the inner function, softmax as normalization, and the outer kernel function combined with Out-projection as the outer function.
        }
        \label{fig:attn-encoder}
\end{figure}

As depicted in Figure~\ref{fig:attn-encoder}, our encoder maintains the ViT structure while replacing standard attention blocks with our proposed MHSA variants. The key modification lies in the multi-head kernel self-attention mechanism, which comprises the In-Projection step, the multi-head kernel self-attention as the inner function, softmax normalization, and the outer kernel function combined with out-Projection as the outer function.

\phantomsection\label{linear block}
In addition to the pseudo-MHSA Block and Gaussian-MHSA Block, we also introduce a Linear-MHSA Block. This block sets the output dimension $H$ of the inner function parameters $\mathbf{W}^{\text{inner}}$ (mentioned in Lemma~\ref{lemma:KA-Kernel}) to match the sequence length $S$. Through this dimensional alignment, the inner function $\Psi$ produces output with identical dimensions to standard attention maps, thereby generating a simulated attention map.\\

\begin{figure}[H]
   \centering
        \centering
        \includegraphics[width=0.8\linewidth]{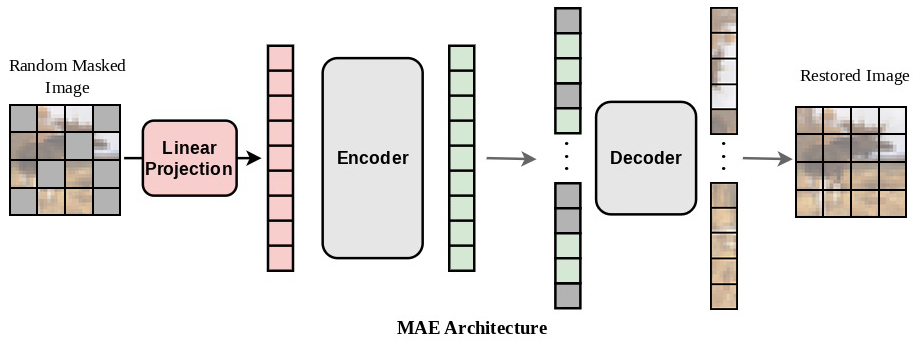}
        \caption{MAE Autoencoder. The basic framework of our model follows the original MAE design as closely as possible, but we replace the original ViT encoder and decoder with our model encoder and decoder. In the linear projection layer, a convolutional layer is employed to map image patches into sequence embeddings. The last layer of the decoder utilizes a transpose convolutional layer to directly reconstruct image patches from the decoder's output sequence embeddings.}
        \label{fig:mae}
\end{figure}

We integrate our encoder into the MAE framework, as shown in Figure~\ref{fig:mae}. The basic framework follows the original MAE design as closely as possible, but we replace the original ViT encoder and decoder with our model encoder and decoder. In the linear projection layer, a convolutional layer is employed to map image patches into sequence embeddings. The last layer of the decoder utilizes a transpose convolutional layer to directly reconstruct image patches from the decoder's output sequence embeddings.

\section{Experiments}
\label{experiment}

In this section, we investigate whether the pseudo-MHSA block can perform comparably to the original multi-head self-attention module, and we validate the effectiveness of the implementation of Lemma~\ref{lemma:KA-Kernel} through Algorithm~\ref{alg:gaussian-mhsa} and the linear block. We compare five models (Table~\ref{tab: model information}):

\begin{itemize}
    \item \textbf{Standard}: Baseline ViT model with standard MHSA.
    \item \textbf{Param-Fusion}: Exact implementation of Algorithm~\ref{alg:pseudo-mhsa}.
    \item \textbf{Semi-Fusion}: Implementation of Algorithm~\ref{alg:pseudo-mhsa} with $\widetilde{\mathbf{W}}^v$ and $\mathbf{W}^O$ retained.
    \item \textbf{Linear-Sim}: Linear-MHSA model with the block mentioned in section~\ref{linear block}.
    \item \textbf{Gaussian}: Exact implementation of Algorithm~\ref{alg:gaussian-mhsa}.
\end{itemize}

We pretrain an 8-layer encoder for the Standard, Param-Fusion, and Semi-Fusion models using the MAE framework on CIFAR-10. Note that the MAE decoder and encoder share the same architecture, differing only in a reduced dimension for the decoder. 

During finetuning, we initialize the first 6 layers of the encoder with the pretrained weights. For the Linear-Sim and Gaussian models, we train a 6-layer encoder from scratch. The representations of the class tokens extracted from the 6-layer encoder are fed into a simple linear classifier. All experiments are implemented using the PyTorch library \cite{PyTorch}, and the ViT encoder model is based on its official implementation. The source code has been uploaded to GitHub, and training hyperparameters are detailed in the appendix~\ref{app:params}.

\subsection{Analysis of Experimental Results}
\begin{figure}[H]
   \centering
        \centering
        \includegraphics[width=0.8\linewidth]{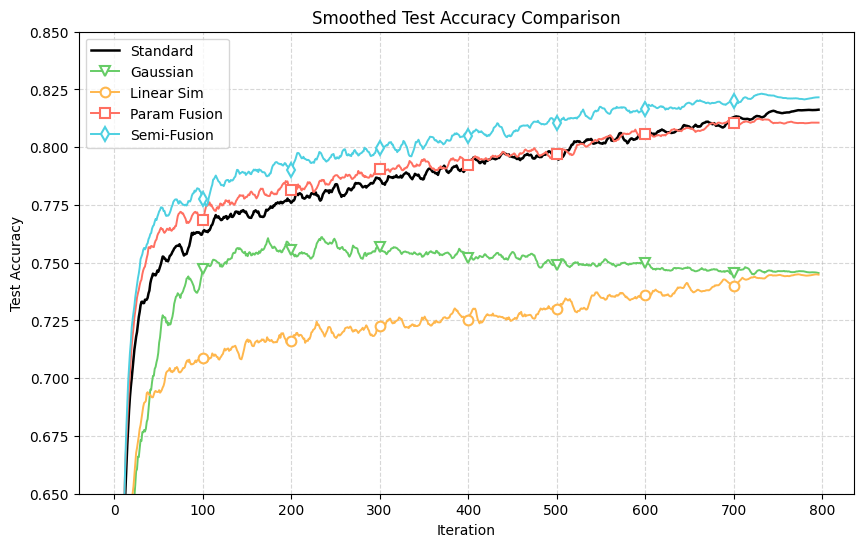}
        \caption{Test accuracy on CIFAR-10.}
        \label{fig: acc}
\end{figure}

As shown in Figure~\ref{fig: acc}, the Semi-Fusion model achieves the highest test accuracy (\textbf{0.8243}), outperforming the Standard model (0.8162), which validates the effectiveness of the low-rank approximation. Param-Fusion (0.8144) maintains performance close to the baseline. Linear-Sim and Gaussian models show lower accuracies (0.7454 and 0.7655), yet Gaussian’s lightweight design (256K params) offers potential for resource-constrained scenarios.

\begin{table}[H]
    \centering
        \begin{tabular}{lccccr}
            \toprule
                Model & $D_{\text{model}}$ &  \# of params & accuracy \\
                \midrule
                Standard        & 256   & 3.20M & 0.8162 & \\
                Param-Fusion    & 256   & 2.45M & 0.8144 & \\
                Semi-Fusion     & 256   & 2.50M & \textbf{0.8243} & \\
                Linear-Sim      & 256   & 5.60M & 0.7454  & \\
                Gaussian        & 64    & \textbf{256K}  & 0.7655 & \\
                \bottomrule
        \end{tabular}
        \caption{Basic information of models.}
        \label{tab: model information}
\end{table}

\subsubsection{Inter-layer Attention Distribution}
\begin{figure}[H]
    \centering
    \includegraphics[width=0.8\linewidth]{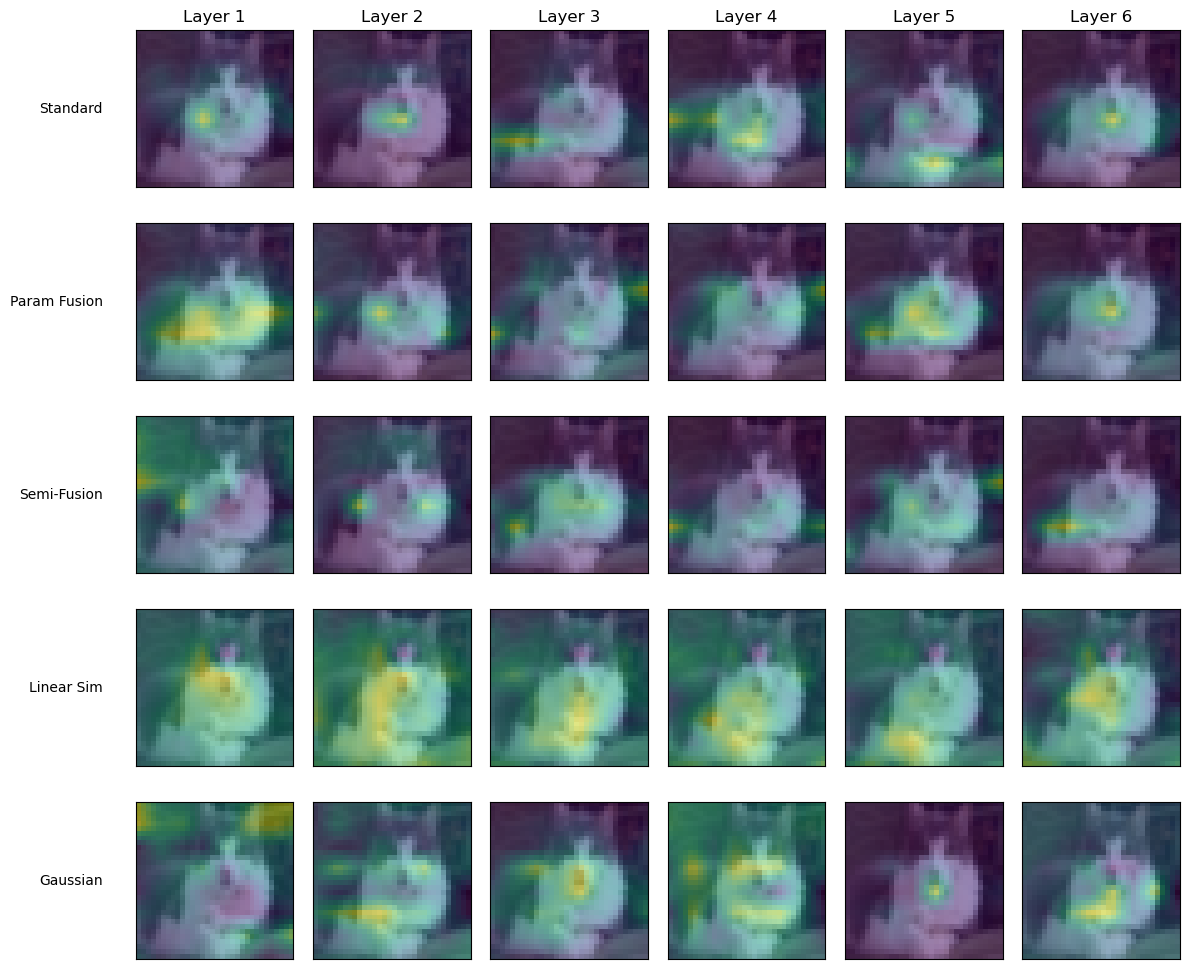}
    \caption{Attention heatmaps of different models with head-averaged results across Layer 1 to Layer 6 in the encoder, illustrating the attention response patterns of each model at various layers.}
    \label{fig:head-mix}
\end{figure}

From Figure~\ref{fig:head-mix}, in bottom layers (Layer 1–2), Standard, Semi-Fusion, and Param-Fusion focus on target outlines and basic features; Linear-Sim scatters attention, while Gaussian highlights local parts. In middle layers (Layer 3–4), the first three models concentrate on the target core, Linear-Sim reduces background attention, and Gaussian balances local/surrounding features. In top layers (Layer 5–6), Semi-Fusion and Param-Fusion retain Standard’s high-level semantic capture; Linear-Sim has a single focus pattern, and Gaussian shows diverse local activations.\\

\subsubsection{Multi-head Attention Patterns}
Attention heatmaps between head is shown in Appendix. The Standard model’s heads differentiate in focusing on subject parts and background. Semi-Fusion and Param-Fusion preserve similar multi-head patterns, retaining feature separation. Linear-Sim’s heads show global coverage due to the linear kernel, while Gaussian’s heads, with the Gaussian kernel, concentrate on diverse local subject parts, demonstrating distinct local similarity capture.\\

In conclusion, the experiment validates the method’s effectiveness in performance, parameter efficiency, and attention mechanism retention. Semi-Fusion excels in accuracy with fewer parameters; Gaussian’s lightweight design provides valuable direction. These model variants lay a solid foundation for future research.\\

\section{Related Works}
\label{sec:related}
\subsection{Kolmogorov-Arnold Networks and Function Approximation}
The theoretical foundation of this work stems from the Kolmogorov-Arnold representation theorem \cite{kolmogorov1957}, which provides a constructive decomposition framework for multivariate continuous functions. Liu, Z. et al\cite{liu2025kankolmogorovarnoldnetworks} proposed Kolmogorov-Arnold Networks, which implement the inner and outer functions of the theorem using learnable B-spline basis functions. This contrasts sharply with traditional deep networks that employ fixed activation functions such as ReLU \cite{Nair2010} or GELU \cite{Hendrycks2016}. We extend the theoretical framework of KANs by replacing spline bases with generalized kernel functions.

\subsection{Self-Attention and Kernel Methods}
The kernel-based interpretation of self-attention mechanisms builds upon the pioneering work of Tsai, Y. H. et al \cite{tsai2019transformerdissectionunifiedunderstanding}, who first formalized the connection between dot-product attention and kernel similarity measures. Our convolutional reinterpretation of self-attention elaborates on the findings of Cordonnier, J. et al \cite{cordonnier2020relationshipselfattentionconvolutionallayers}, who demonstrated the equivalence between self-attention layers and dynamic convolutions.

\subsection{Efficient Transformer Architectures}
The low-rank approximation strategy proposed in this work offers new insights into efficient Transformer variants. The parameter reduction mechanism is related to the low-rank projection method of Linformer \cite{wang2020linformerselfattentionlinearcomplexity} to some extent, while the kernel-based implementation shares philosophical similarities with the orthogonal random features of Performer \cite{choromanski2022rethinkingattentionperformers}. However, our method only decomposes and combines dimensions without performing matrix eigen decomposition.

\section{Conclusions}
\label{sec:conclusions}
In this paper, we propose a unified kernel method framework under which we reinterpret KANs and self-attention mechanisms through the lens of kernel expansions, revealing a theoretical connection between these two paradigms in function approximation. The proposed Pseudo-MHSA module demonstrates that a low-rank decomposition from a multi-head perspective can reduce the parameter count by 23\% while maintaining competitive performance on the CIFAR-10 classification task. Meanwhile, the Gaussian-MHSA variant further validates the feasibility of non-linear kernel attention mechanisms, with accuracy loss kept within acceptable limits.\\

Future research should address two key limitations of the current work: (1) the experiments are limited to medium-scale vision tasks, and their generalizability needs to be verified on large-scale multimodal datasets; (2) the explicit kernel tensors occupy a significant amount of GPU memory, posing challenges for practical applications and necessitating optimizations at the CUDA kernel level \cite{NVIDIA_CUDA}.\\

Promising directions for extension include hybrid architectures that combine linear and non-linear kernels across network depths, dynamic kernel selection via gating mechanisms, and the application of this method to operator learning tasks. In deeper layers of neural networks or during the inference phase, the use of trainable reference matrices or fixed feature vectors may facilitate the compression of attention matrices.

\newpage
\bibliographystyle{unsrt}  
\bibliography{PRIMEarxiv}  

\begin{thebibliography}{10}

\bibitem{LeCun2015}
Yann LeCun, Yoshua Bengio, and Geoffrey Hinton.
\newblock Deep learning.
\newblock {\em Nature}, 521(7553):436--444, 2015.

\bibitem{liu2025kankolmogorovarnoldnetworks}
Ziming Liu, Yixuan Wang, Sachin Vaidya, Fabian Ruehle, James Halverson, Marin Soljačić, Thomas~Y. Hou, and Max Tegmark.
\newblock Kan: Kolmogorov-arnold networks, 2025.

\bibitem{kolmogorov1957}
A.~N. Kolmogorov.
\newblock On the representation of continuous functions of many variables by superposition of continuous functions of one variable and addition.
\newblock {\em Doklady Akademii Nauk SSSR}, 114(5):953--956, 1957.

\bibitem{Nair2010}
Vinod Nair and Geoffrey~E. Hinton.
\newblock Rectified linear units improve restricted boltzmann machines.
\newblock In {\em Proceedings of the 27th International Conference on Machine Learning (ICML)}, pages 807--814, 2010.

\bibitem{Britannica2023}
Sigmoid function.
\newblock {\em Encyclopaedia Britannica}, 2023.

\bibitem{Vaswani2017}
Ashish Vaswani, Noam Shazeer, Niki Parmar, Jakob Uszkoreit, Llion Jones, Aidan~N. Gomez, Lukasz Kaiser, and Illia Polosukhin.
\newblock Attention is all you need.
\newblock In {\em Advances in Neural Information Processing Systems (NeurIPS)}, pages 5998--6008, 2017.

\bibitem{deBoor1972}
Carl de~Boor.
\newblock On calculating with b-splines.
\newblock {\em Journal of Approximation Theory}, 6:50--62, 1972.

\bibitem{Krizhevsky2009}
Alex Krizhevsky.
\newblock Learning multiple layers of features from tiny images.
\newblock Technical report, University of Toronto, 2009.

\bibitem{Dosovitskiy2020}
Alexey Dosovitskiy, Lucas Beyer, Alexander Kolesnikov, Dirk Weissenborn, Xiaohua Zhai, Thomas Unterthiner, Mostafa Dehghani, Matthias Minderer, Georg Heigold, Sylvain Gelly, Jakob Uszkoreit, and Neil Houlsby.
\newblock An image is worth 16x16 words: Transformers for image recognition at scale.
\newblock {\em arXiv preprint arXiv:2010.11929}, 2020.

\bibitem{He2021}
Kaiming He, Xinlei Chen, Saining Xie, Yanghao Li, Piotr Dollár, and Ross Girshick.
\newblock Masked autoencoders are scalable vision learners.
\newblock {\em arXiv preprint arXiv:2111.06377}, 2021.

\bibitem{zhang2008}
Xinjian Zhang and Han Long.
\newblock {\em Spline Functions and Reproducing Kernels}.
\newblock National University of Defense Technology Press, Changsha, 2008.

\bibitem{cordonnier2020relationshipselfattentionconvolutionallayers}
Jean-Baptiste Cordonnier, Andreas Loukas, and Martin Jaggi.
\newblock On the relationship between self-attention and convolutional layers, 2020.

\bibitem{Golub1996}
Gene~H. Golub and Charles F.~Van Loan.
\newblock {\em Matrix Computations}.
\newblock Johns Hopkins University Press, 3rd edition, 1996.

\bibitem{Scholkopf2002}
Bernhard Sch{\"o}lkopf and Alexander~J. Smola.
\newblock {\em Learning with Kernels: Support Vector Machines, Regularization, Optimization, and Beyond}.
\newblock MIT Press, 2002.

\bibitem{Ba2016}
Jimmy~Lei Ba, Jamie~Ryan Kiros, and Geoffrey~E. Hinton.
\newblock Layer normalization.
\newblock {\em arXiv preprint arXiv:1607.06450}, 2016.

\bibitem{PyTorch}
{PyTorch Contributors}.
\newblock {PyTorch: An Open Source Machine Learning Framework}, 2025.

\bibitem{Hendrycks2016}
Dan Hendrycks and Kevin Gimpel.
\newblock Gaussian error linear units (gelus).
\newblock {\em arXiv preprint arXiv:1606.08415}, 2016.

\bibitem{tsai2019transformerdissectionunifiedunderstanding}
Yao-Hung~Hubert Tsai, Shaojie Bai, Makoto Yamada, Louis-Philippe Morency, and Ruslan Salakhutdinov.
\newblock Transformer dissection: A unified understanding of transformer's attention via the lens of kernel, 2019.

\bibitem{wang2020linformerselfattentionlinearcomplexity}
Sinong Wang, Belinda~Z. Li, Madian Khabsa, Han Fang, and Hao Ma.
\newblock Linformer: Self-attention with linear complexity, 2020.

\bibitem{choromanski2022rethinkingattentionperformers}
Krzysztof Choromanski, Valerii Likhosherstov, David Dohan, Xingyou Song, Andreea Gane, Tamas Sarlos, Peter Hawkins, Jared Davis, Afroz Mohiuddin, Lukasz Kaiser, David Belanger, Lucy Colwell, and Adrian Weller.
\newblock Rethinking attention with performers, 2022.

\bibitem{NVIDIA_CUDA}
{NVIDIA Corporation}.
\newblock {\em {CUDA} C++ Programming Guide}, 2023.

\end{thebibliography}

\newpage
\appendix
\section{ATTENTION HEATMAPS FOR EACH HEAD}
\label{app:heatmapheads}
\begin{figure}[H]
    \centering
    \includegraphics[width=0.9\linewidth]{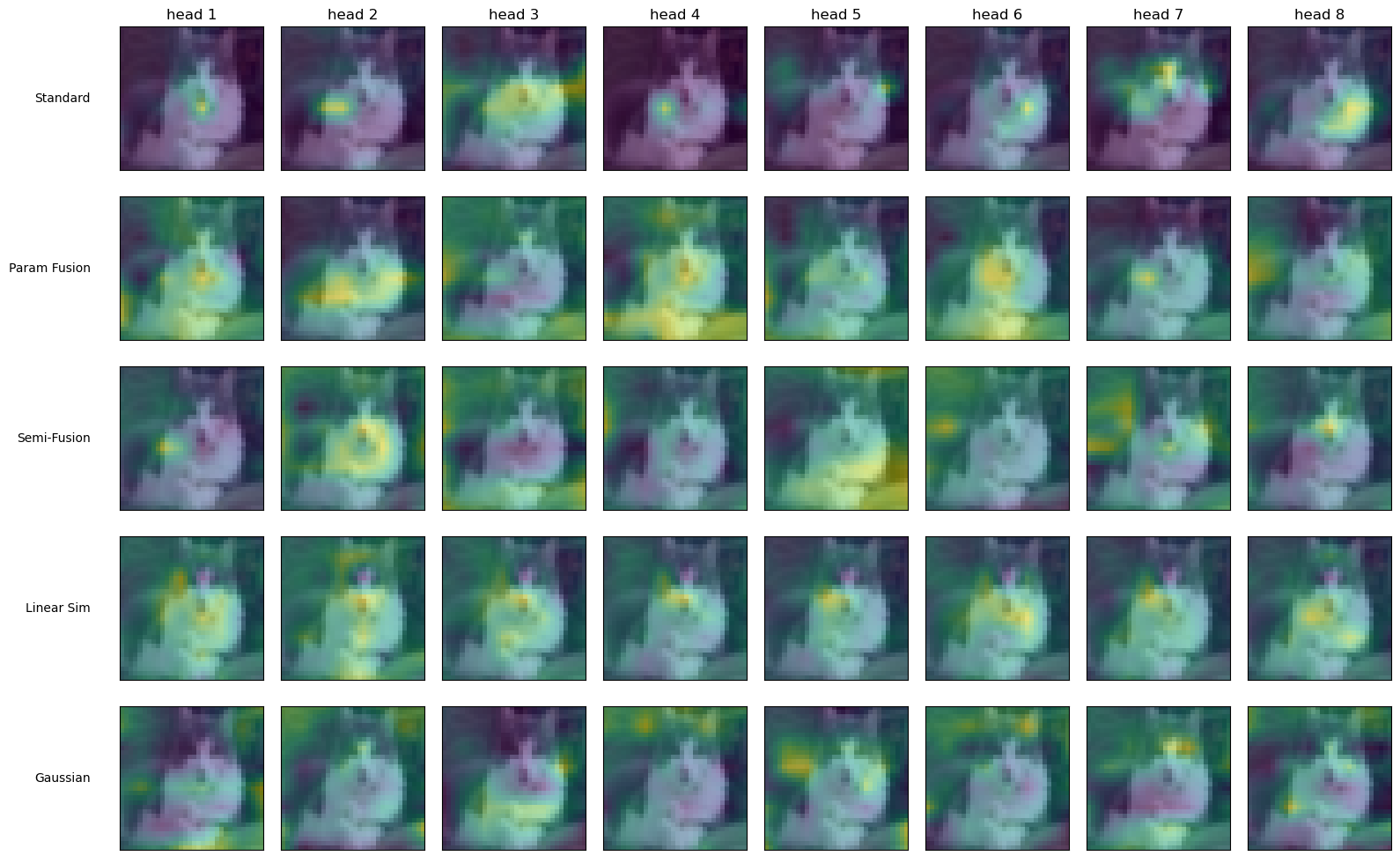}
    \caption{Attention heatmaps for each head of different models at layer 1.}
    \label{fig:layer1}
\end{figure}

\begin{figure}[H]
    \centering
    \includegraphics[width=0.9\linewidth]{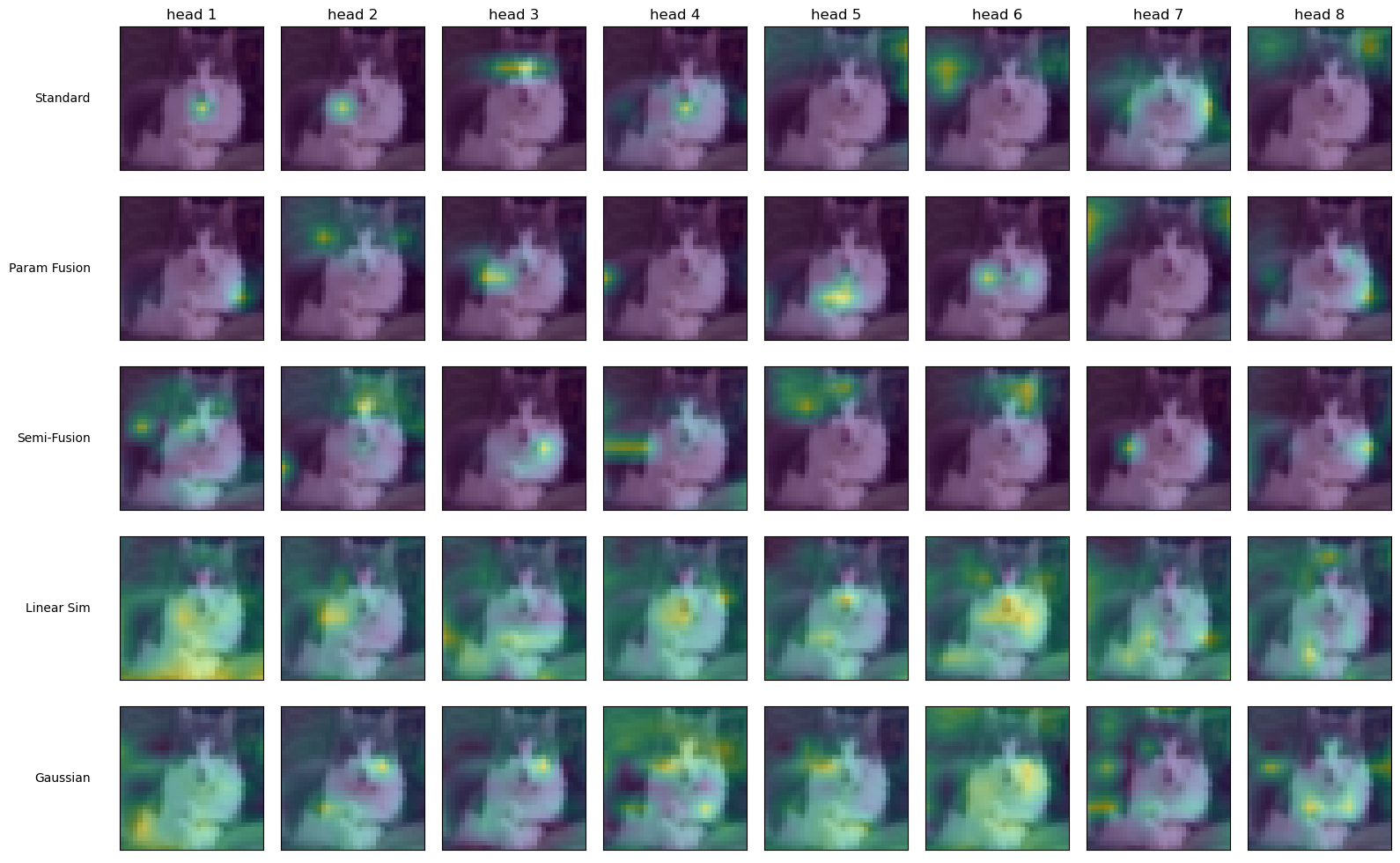}
    \caption{Attention heatmaps for each head of different models at layer 2.}
    \label{fig:layer2}
\end{figure}

\begin{figure}[htbp]
    \centering
    \includegraphics[width=0.9\linewidth]{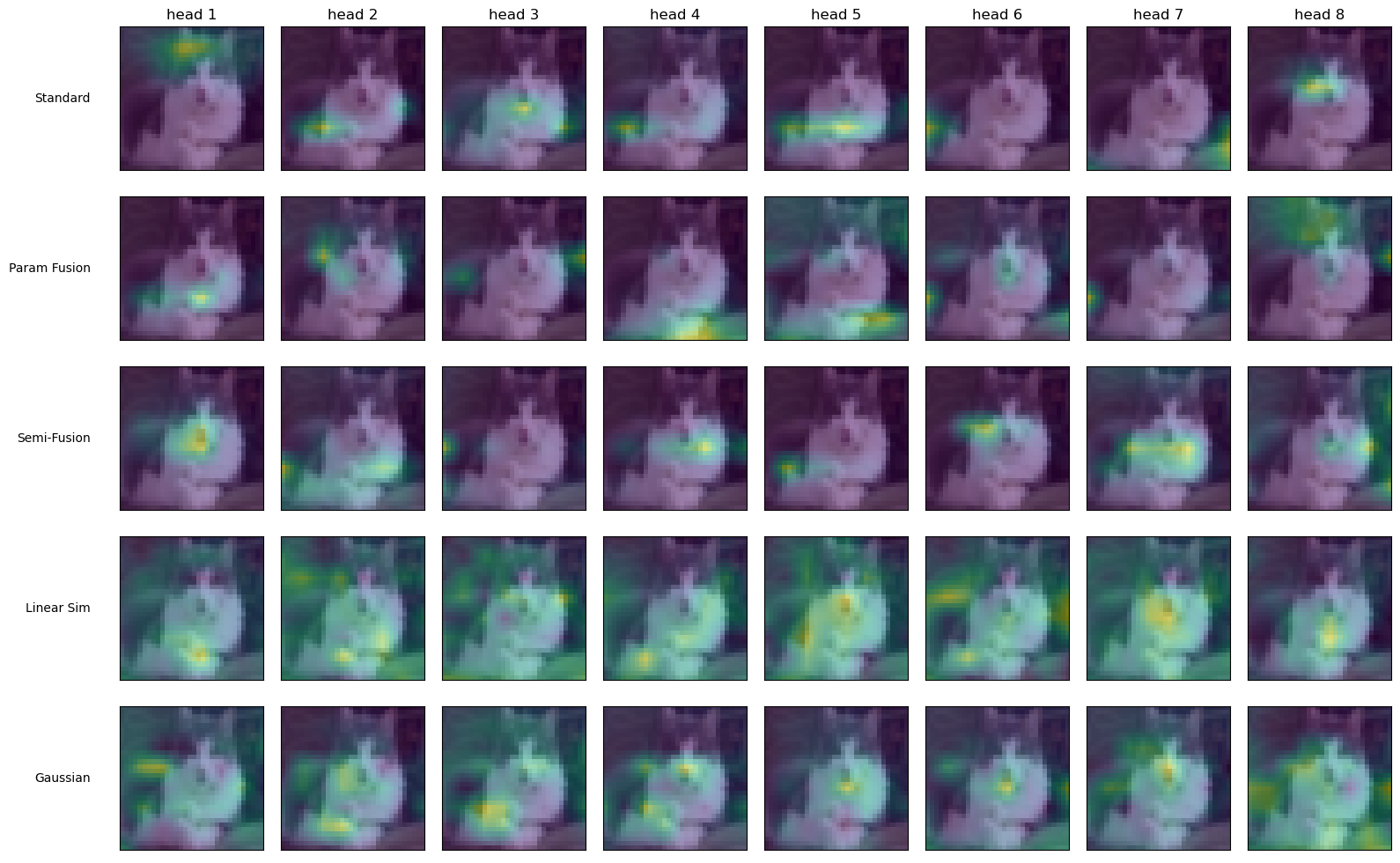}
    \caption{Attention heatmaps for each head of different models at layer 3.}
    \label{fig:layer3}
\end{figure}

\begin{figure}[htbp]
    \centering
    \includegraphics[width=0.9\linewidth]{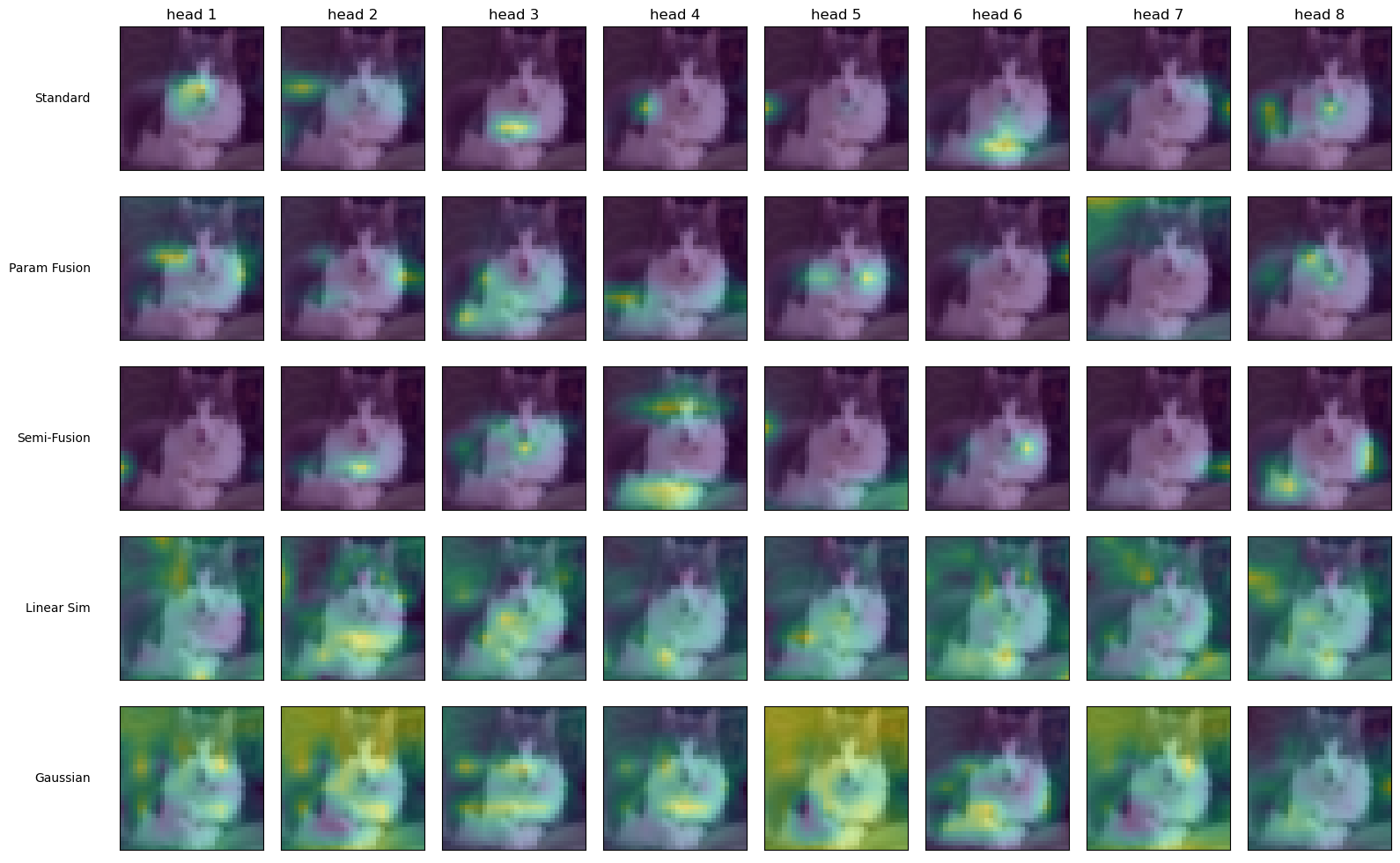}
    \caption{Attention heatmaps for each head of different models at layer 4.}
    \label{fig:layer4}
\end{figure}

\begin{figure}[htbp]
    \centering
    \includegraphics[width=0.9\linewidth]{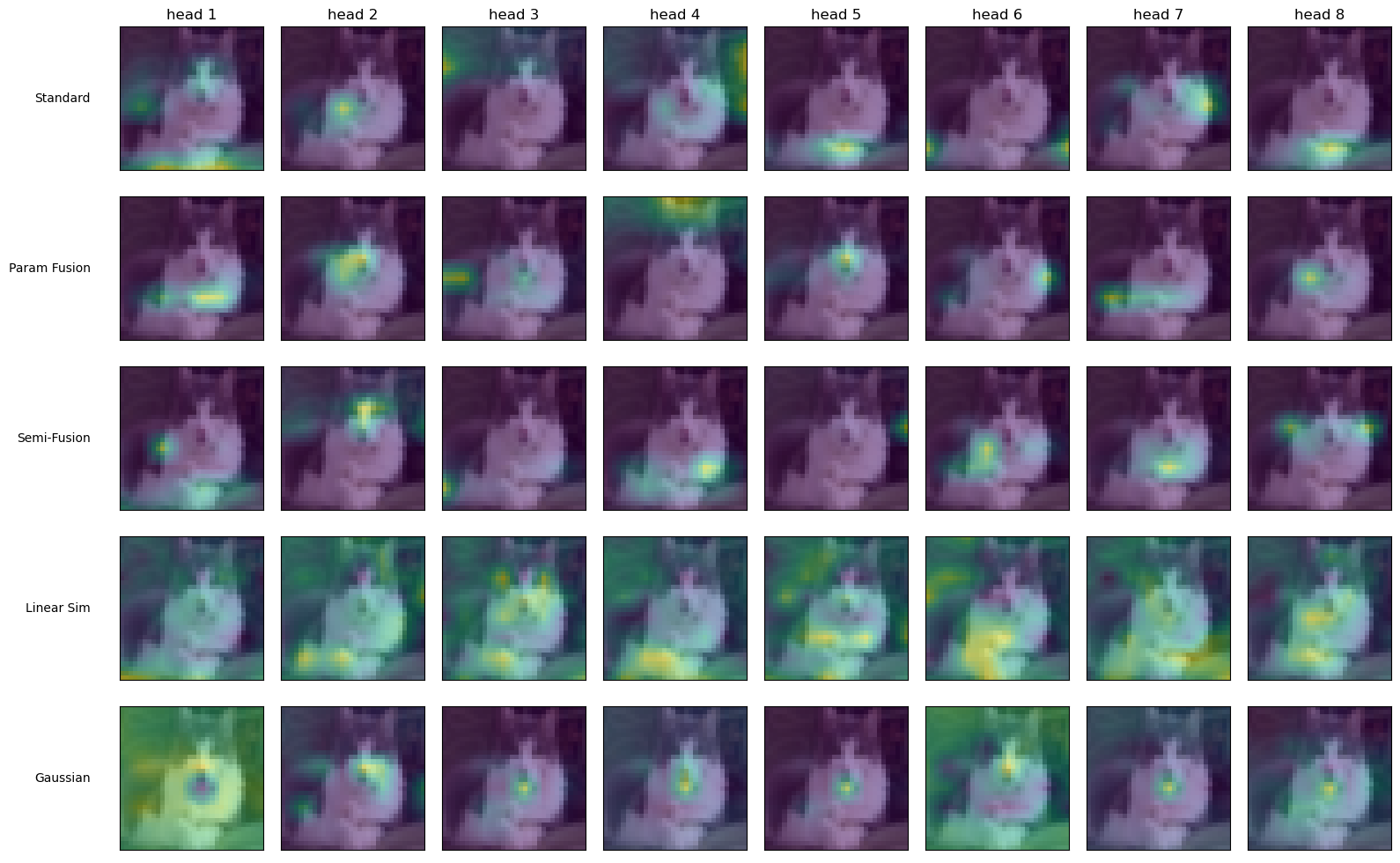}
    \caption{Attention heatmaps for each head of different models at layer 5.}
    \label{fig:layer5}
\end{figure}

\begin{figure}[htbp]
    \centering
    \includegraphics[width=0.9\linewidth]{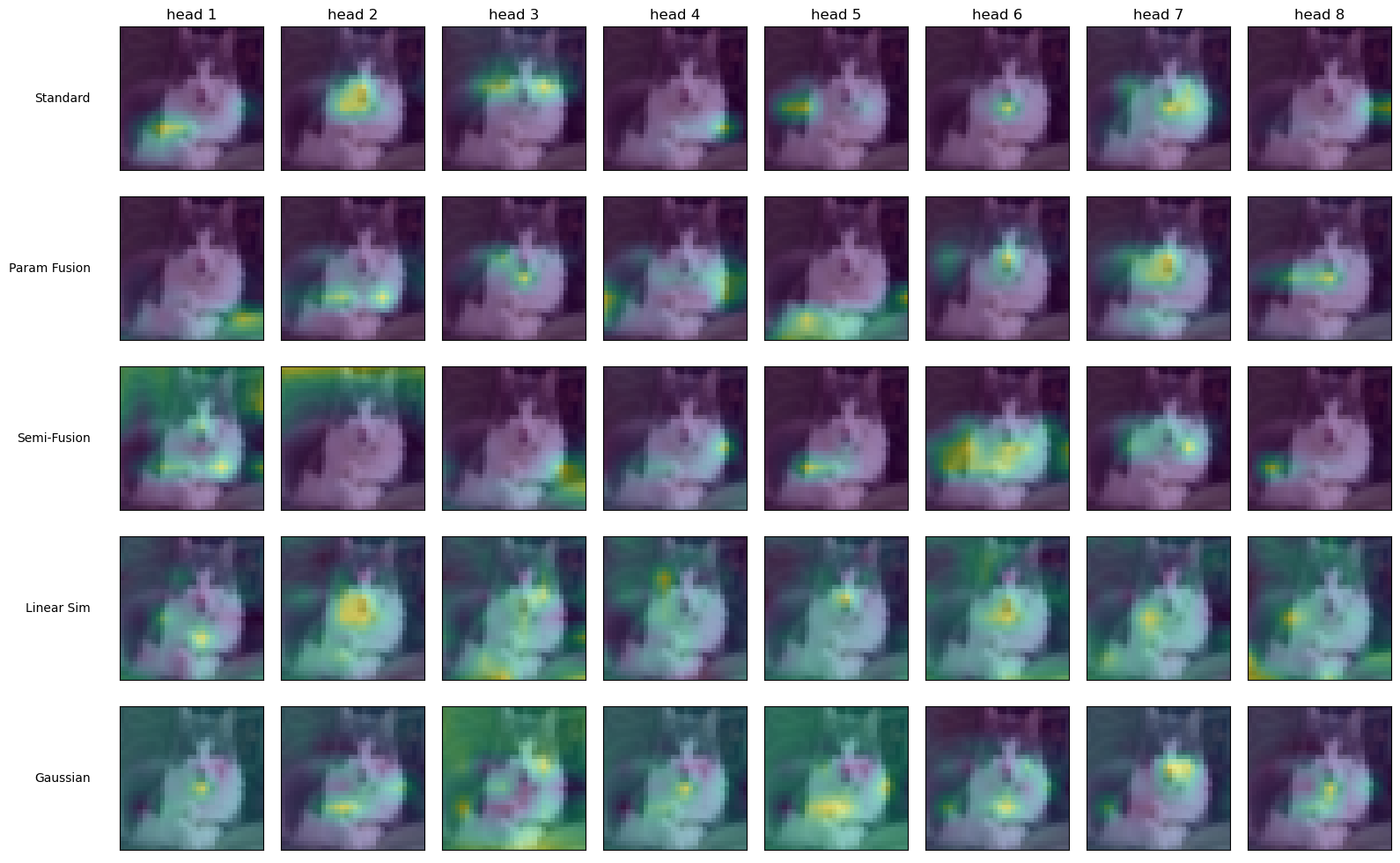}
    \caption{Attention heatmaps for each head of different models at layer 6.}
    \label{fig:layer6}
\end{figure}

\newpage
\section{HYPER-PARAMETERS USED IN OUR EXPERIMENTS}
\label{app:params}
\begin{table}[htbp]
\centering
\begin{minipage}[t]{0.48\textwidth} 
\centering
\begin{tabular}{lr}
\toprule
\textbf{Parameters} & \textbf{Value} \\
\midrule
encoder dimension & 256 \\
encoder mlp dimension & 512 \\
encoder layers & 8 \\
decoder dimension & 192 \\
decoder mlp dimension & 384 \\  
decoder layers & 6 \\            
number of heads & 8 \\
attention dropout rate & 0 \\
mlp dropout rate & 0.1 \\
dropout rate & 0.1 \\
\midrule
learning rate & $10^{-3}$ \\
warmup ratio & 0.05 \\
betas ($\beta_1, \beta_2$) & (0.9, 0.95) \\
weight decay & $10^{-4}$ \\
batch size & 100 \\
training epochs & 1600 \\
AdamW & $\checkmark$ \\
cosine decay & $\checkmark$ \\
\bottomrule
\end{tabular}
\label{tab:pretrain}
\end{minipage}
\hfill
\begin{minipage}[t]{0.48\textwidth} 
\centering
\begin{tabular}{lr}
\toprule
\textbf{Parameters} & \textbf{Value} \\
\midrule
encoder dimension & 256 \\
encoder mlp dimension & 512 \\
encoder layers & 6 \\
number of heads & 8 \\
attention dropout rate & 0 \\
mlp dropout rate & 0.1 \\
dropout rate & 0.1 \\
\midrule
learning rate & $10^{-3}$ \\
warmup ratio & 0.05 \\
betas ($\beta_1, \beta_2$) & (0.9, 0.95) \\
weight decay & $10^{-4}$ \\
batch size & 100 \\
training epochs & 800 \\
AdamW & $\checkmark$ \\
cosine decay & $\checkmark$ \\
\bottomrule
\end{tabular}
\label{tab:finetune}
\end{minipage}
\caption{Pretrain parameters and finetune parameters.}
\end{table}

\end{document}